\documentclass[runningheads, envcountsame, a4paper]{llncs}
\usepackage{mathtools}
\usepackage[noend]{algpseudocode}
\usepackage{booktabs}
\usepackage{subcaption}
\usepackage{numprint}
\usepackage{soul}
\usepackage[hyphens]{url}
\usepackage{graphicx}
\usepackage{booktabs}
\usepackage{bbding}

\usepackage{amsthm}

\usepackage{amsmath,amssymb,amsfonts}
\usepackage{textcomp}
\usepackage{todonotes}
\usepackage{algorithm2e}
\usepackage{scrextend}
\usepackage{tabularx}
\usepackage{xcolor}
\usepackage{multirow}
\usepackage{blindtext}
\usepackage{subcaption}
\usepackage[misc]{ifsym}

\theoremstyle{definition}
\usepackage{amsmath}

\newcommand{\spotlight}{\textsc{SpotLight}}
\newcommand{\anomalous}{\textsc{Anomalous}}
\newcommand{\dominant}{\textsc{Dominant}}
\newcommand{\sskn}{\textsc{SSKN}}
\newcommand{\amen}{\textsc{AMEN}}

\newcommand{\acm}{\textsc{ACM}}

\newcommand{\blogc}{\textsc{BlogC}}
\newcommand{\darpa}{\textsc{DARPA}}
\newcommand{\enron}{\textsc{Enron}}

\newcommand{\system}{{GraphAnoGAN}}

\begin{document}
	\title{\system: Detecting Anomalous Snapshots from Attributed Graphs}
	\toctitle{\system: Detecting Anomalous Snapshots from Attributed Graphs}
	\tocauthor{Siddharth~Bhatia, Yiwei~Wang, Bryan~Hooi, Tanmoy~Chakraborty}
	%
	%
	
		\author{
			Siddharth Bhatia(\Letter)\inst{1} \and
			Yiwei Wang\inst{1} \and
			Bryan Hooi\inst{1} \and
			Tanmoy Chakraborty\inst{2}
		}
		\authorrunning{S. Bhatia et al.}
		%
		\institute{
			National University of Singapore \\
			\email{\{siddharth, y-wang, bhooi\}@comp.nus.edu.sg} \and
			IIIT-Delhi, India \\
			\email{tanmoy@iiitd.ac.in}
		}

	\maketitle              
	\begin{abstract}
		Finding anomalous snapshots from a graph has garnered huge attention recently. Existing studies address the problem using shallow learning mechanisms such as subspace selection, ego-network, or community analysis. These models do not take into account the multifaceted interactions between the structure and attributes in the network. In this paper, we propose \system, an anomalous snapshot ranking framework, which consists of two core components -- generative and discriminative models. Specifically, the generative model learns to approximate the distribution of anomalous samples from the candidate set of graph snapshots, and the discriminative model detects whether the sampled snapshot is from the ground-truth or not. Experiments on $4$ real-world networks show that \system\ outperforms $6$ baselines with a significant margin ($28.29\%$ and $22.01\%$ higher precision and recall, respectively compared to the best baseline, averaged across all datasets).
	\end{abstract}

\section{Introduction}
Anomaly detection on graphs is a well-researched problem and plays a critical role in cybersecurity, especially network security \cite{borghesi2019anomaly}. Majority of the proposed approaches focus on anomalous nodes \cite{kleinberg1999authoritative,akoglu2010oddball,jiang2016catching,anomalous}, anomalous edges \cite{Sricharan,sskn,sedanspot}, community structures \cite{sun2007graphscope}, or sudden surprising changes in graphs \cite{bhatia2020midas,belth2020mining,chang2021f}.

However, we focus our attention on \textbf{\textit{detecting anomalous snapshots from attributed graphs}}. This problem is motivated by the following cybersecurity threats: (a) fraudulent customers controlling the sentiment (customers operate in a way that they can not be tracked individually), (b) hackers targeting the network (attacks such as DDOS, phishing), (c) black-market syndicates in online social media \cite{dutta2020blackmarket}, and (d) camouflaged financial transactions.

Detecting anomalous snapshots in a graph has received little attention; \spotlight\ \cite{spotlight} is one of them. However, \spotlight\ does not take into account the patterns being formed in the graph even if there is no outburst of edges. Moreover, it tends to ignore the node features as well. On the other hand, convolutional architectures nicely capture the complex interactions between the structure and the attributes, taking data sparsity and non-linearity into account.

Therefore, we propose \textbf{\system}, a  generative adversarial network (GAN) based framework that takes advantage of its structure in the following two ways: (i) the generative model learns to find the anomalous snapshots via the signals from the discriminative model and the global graph topology; (ii) the discriminative model achieves the improved classification of snapshots by modeling the data provided by the generative model and ground-truth. To the best of our knowledge, \system\ is the first GAN-based method for detecting anomalous snapshots in graphs.

Our convolutional architecture plays an integral role in choosing the required feature descriptors, to be utilized in identifying anomalous snapshots. Moreover, we demonstrate that \system\ is able to learn complex structures and typical patterns required in such problem settings.

Finally, we evaluate the performance of \system~on $4$ datasets and compare it with $6$ state-of-the-art baselines. Experimental results show that \system~outperforms all the baselines by a significant margin -- it achieves  $65.75\%$ precision ({\em resp.} $66.5\%$ recall) on average across all the datasets, which is $28.29\%$ ({\em resp.} $22.01\%$) higher than the best baseline.

\paragraph{Reproducibility:} Our code and datasets are publicly available at \url{https://github.com/LCS2-IIITD/GraphAnoGAN-ECMLPKDD21}.

\section{Related Work}\label{sec:relatedwork}

Readers are encouraged to go through \cite{akoglu2015graph} and \cite{Mattia2019ASO} for extensive surveys on graph-based and GAN-based anomaly detection. Traditional methods for anomaly detection can be (i) reconstruction-based: PCA \cite{jolliffe1986principal}, kernel PCA \cite{gunter2007fast}; (ii) clustering-based: GMM \cite{zimek2012survey}, KDE \cite{aggarwal2015outlier}; (iii) one class classification-based: OC-SVM \cite{scholkopf2001estimating}. More recently, deep learning based methods for anomaly detection have been popular. These methods include (i) energy-based: DSEBM \cite{zhai2016deep}, MEG \cite{kumar2019maximum}; (ii) autoencoder-based: DAGMM \cite{zong2018deep}; and (iii) GAN-based: AnoGAN \cite{schlegl2017unsupervised}, Ganomaly \cite{akcay2018ganomaly}, FenceGAN \cite{ngo2019fence}, MemGAN \cite{yang2020memgan}, TAnoGAN \cite{Bashar2020TAnoGANTS}, \cite{Han2020GANEF}, and ExGAN \cite{bhatia2021exgan}. There has been work on attributed graphs as well \cite{Soundarajan2016GeneratingGS,Bendimerad2019SIASminerMS,Pei2020ResGCNAD,Pienta2014MAGEMA,Atzmller2019MinerLSDEM,Perozzi2018DiscoveringCA,Baroni2017EfficientlyCV}. However, these methods do not detect graph anomalies.

\textbf{Graph-based GAN Frameworks:} GraphGAN \cite{graphgan} has two components -- Generator which tries to model the true connectivity distribution for all the vertices, and Discriminator which detects whether the sampled vertex is from the ground-truth or generated by the Generator. NetGAN \cite{netgan} uses a novel LSTM architecture and generates graphs via random walks. Generator in such models tries to form the whole graph which is computationally challenging and not scalable. Instead of generating the whole graph, our proposed Generator in \system\ learns to retrieve anomalous snapshots from a pool via signals from the Discriminator.

\textbf{Graph-Based Anomaly Detection:} These
methods can be divided into four categories. (i) Using community or ego-network analysis to spot the anomaly. \amen~\cite{amen}  detects anomalous neighborhoods in attributed graphs.  A neighborhood is considered normal if it is internally well-connected and nodes are similar to each other on a specific attribute subspace as well as externally well separated from the nodes at the boundary. \spotlight~\cite{spotlight} used a randomized sketching-based approach, where an anomalous snapshot is placed at a larger distance from the normal snapshot. (ii) Utilizing aggregated reconstruction error of structure and attribute. \dominant\: \cite{sdm} has a GCN and autoencoder network, where an anomaly is reported if aggregated error breaches the threshold. (iii) Using residuals of attribute information and its coherence with graph information. \anomalous\: \cite{anomalous} is a joint framework to conduct attribute selection and anomaly detection simultaneously based on cut matrix decomposition \cite{mahoney2009cur} and residual analysis. (iv) Performing anomaly detection on edge streams. \sskn\: \cite{sskn} takes neighbors of a node, and historic edges into account to classify an edge. We consider all of these as baselines.

Since we focus on attributed graphs, the abnormality is determined jointly by mutual interactions of nodes (i.e., topological structure) and their features (i.e., node attributes). As shown in Table \ref{tab:comparison}, \system\ satisfies all the four aspects -- it takes into account {\em node attributes}; it {\em classifies graph snapshots}; it can be {\em generalized} for weighted/unweighted and directed/undirected graphs; and it considers  {\em structures/patterns} exhibited by anomalies.

\begin{table}[!ht]
\centering
\caption{Comparison of \system\ with baseline approaches.}
\label{tab:comparison}
\makebox[\textwidth][c]{
\begin{tabular}{@{}lccccc|c@{}}
\toprule
& {\multirow{1}{*}{\spotlight}} & {\multirow{1}{*}{\amen}} & {\multirow{1}{*}{\anomalous}} & {\multirow{1}{*}{\dominant}} & {\multirow{1}{*}{\sskn}} & {\multirow{1}{*}{\textbf{\system}}}  \\\midrule
\textbf{Node attribute}	& & \checkmark & \checkmark & \checkmark &  & \CheckmarkBold \\		
\textbf{Snapshot anomaly} & \checkmark & \checkmark & & & & \CheckmarkBold \\		
\textbf{Generalizable} & \checkmark & \checkmark & \checkmark & \checkmark & \checkmark & \CheckmarkBold \\	
\textbf{Structure/pattern} & & & & & & \CheckmarkBold \\\bottomrule
\end{tabular}}
\end{table}

\section{Problem Definition}
Let $g=\left\{{g}_{1},\cdots,{g}_{T}\right\}$ be $T$ different snapshots\footnote{A `snapshot' of a graph is used as a general term and can refer to any subgraph of the graph, e.g., a particular area of the graph, a temporal snapshot of the graph, an egonet of a node, etc.} of an attributed graph ${G}=\{{V}, {E}, {X}\}$ with $|{V}|=n$ nodes, $|{E}|=m$ edges, and $|{X}|=d$ node attributes. 
Each snapshot $g_t=\{V_t,E_t,X_t\}$ contains $|{V_t}|=n_t$ nodes, $|{E_t}|=m_t$ edges, and each node in $V_t$ is associated with a $d$-dimensional attribute vector $X_t$. $A$ and $A_t$ indicate the adjacency matrices of $G$ and $g_t$, respectively. Each graph snapshot ${g}_{t}$ is associated with a label, $y_{t} \in \mathcal{Y}$, where $\mathcal{Y} \in \{0, 1\}$ ($0$ represents  normal and $1$ represents anomalous snapshot). Our goal is to detect anomalous snapshots by leveraging node attributes, structure, and complex interactions between different information modalities. The aim is to learn a model which could utilise the above information, analyse and identify snapshots that are anomalous in behaviour.

The problem of detecting anomalous graph snapshots from attributed graphs is as follows:  
Given a set of snapshots $\left\{{g}_{1},\cdots,{g}_{T}\right\}$ from a graph ${G}=\{{V}, {E}, {X}\}$ with node attributes, analyze the structure and attributes of every snapshot, and return the top $K$ anomalous snapshots.

\section{Proposed Algorithm}
\begin{figure}[!ht]
    \centering
    \includegraphics[width=\columnwidth]{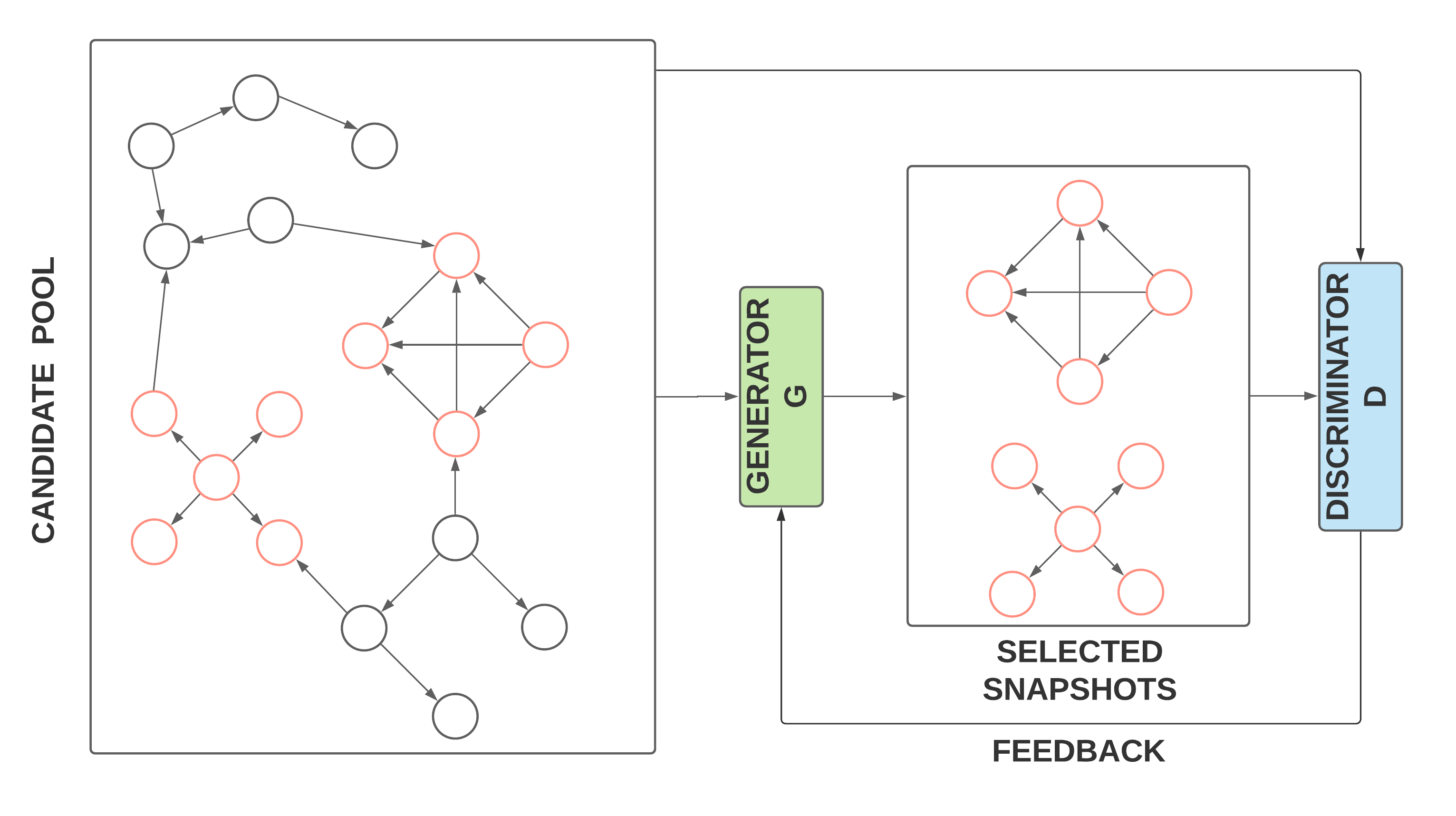}
    \caption{Illustration of \system. The aim of Generator is to report $K$ anomalous snapshots from a pool of samples. Discriminator is fed with samples, of which it identifies if the sample belongs to the ground-truth or is produced by Generator.}
    \label{fig:architecture}
\end{figure}

We introduce \system\ to detect anomalous snapshots of a given graph. \system\ captures complex structures and unusual patterns to rank the snapshots according to the extent of their anomalous behavior. \system\ follows a typical GAN architecture. There are two components: a Generator and a Classifier/Discriminator. The Generator will select those snapshots from the candidate pool which it deems to be anomalous (similar to the ground-truth), and therefore, fools the Discriminator; whereas the Discriminator will distinguish between the ground-truth and the generated snapshots. Essence of our architecture is that the Discriminator will try to classify the graph snapshots, and in doing so, learn the representation of anomalous/normal snapshots. On the other hand, the Generator will learn to find a list of anomalous snapshots from the candidate set. Figure \ref{fig:architecture} depicts the schematic architecture of \system. 

\subsection{GAN modeling}
Given the set of $T$ candidate snapshots $\left\{{g}_{1},\cdots,{g}_{T}\right\}$ of $G$, we want to detect $k$ anomalous snapshots where $k \ll T$. 
We unify two different types of models (i.e., Generator and Discriminator) through a minimax game. The formulation is described  below \cite{irgan}:

\begin{equation}\label{eq:obj}
\begin{array}{c}{J}=\min _{\theta} \max _{\phi} 
\Big(\mathbb{E}_{g_t \sim p_{\mathrm{true}}(g)}[\log D(g_t)]+  {\mathbb{E}_{g_t \sim p_{\theta}(g)}[\log (1-D(g_t))]\Big)}\end{array}
\end{equation}

Here we represent parameters for Generator and Discriminator as $\theta$ and $\phi$, respectively. $p_{true}(g)$ represents the distribution of anomalous snapshots in the ground-truth. Generator and Discriminator are written as $p_{\theta}(g)$ and $f_{\phi}(g_t)$, respectively. 
Discriminator score, $D$ captures the probability of the snapshot being sampled from the ground-truth, calculated using sigmoid function represented below: 

\begin{equation} \label{eq:sig}\small
D(g_t)=\sigma(f_{\phi}(g_t))=\frac{\exp (f_{\phi}(g_t))}{1+\exp(f_{\phi}(g_t))}
\end{equation}

Discriminator $f_{\phi}(g_t)$ is trained on the labeled snapshots and instances received from the Generator. The objective of Discriminator is to maximize the log-likelihood of correctly distinguishing anomalous snapshots from the ones provided by Generator. 

Generator $p_{\theta}(g)$ tries to generate (or select) anomalous snapshots from the candidate pool, i.e., approximates $p_{true}(g)$ as much as possible. Generator learns the distribution of the anomalous snapshots using the information of the entire graph and relative placement of the snapshots within it. Explicit architecture and details are mentioned in Section \ref{sec:architecturesection}. Equation \ref{eq:obj} shows that the optimal parameters of Generator and Discriminator, which  are learned in a way such that the Generator minimizes the objective function, while the Discriminator maximizes the objective function.

Our GAN structure is heavily inspired by IRGAN \cite{irgan}. We here adopt it for the problem of detecting anomalous graphs, and describe the learning algorithm for application in this context.

\paragraph{Discriminator optimization:} 

With the observed anomalous snapshots, and the ones sampled from the current optimal Generator $p_{\theta}(g)$, we obtain the optimal parameters for the Discriminator:

\begin{equation} \label{eq:discri}\small
\begin{array}{c}{\phi^{*}}=\arg \max _{\phi} (\mathbb{E}_{g_t \sim p_{\mathrm{true}}(g)}[\log (\sigma(f_{\phi}{(g_t))]+}  {\mathbb{E}_{g_t \sim p_{\theta^{*}}(g)}[\log (1-\sigma(f_{\phi}(g_t))] \big)}\end{array} 
\end{equation}

\paragraph{Generator optimization:}
Generator minimizes the following objective function to obtain optimal parameters for the model: 
\begin{equation} \label{eq:gen}\small
\theta^{*}=\arg \min _{\theta} \mathbb{E}_{g_t \sim p_{\mathrm{true}} (g)} [\log \sigma (f_{\phi}(g_t))] 
+\mathbb{E}_{g_t \sim p_{\theta}(g)}[\log (1-\sigma(f_{\phi}(g_t)))]
\end{equation} 

Taking reference of Equation \ref{eq:sig}, Equation \ref{eq:gen} can be rewritten as:
\begin{equation} \label{eq:gen2}\small
\begin{split}
\theta^{*}&=\arg \min _{\theta} \mathbb{E}_{g_t \sim p_{\theta}(g)}[\log \frac{\exp (f_{\phi}(g_t))}{(1+\exp (f_{\phi}(g_t)))^2}] \\
&\simeq \arg \max _{\theta} \underbrace{\mathbb{E}_{g_t \sim p_{\theta}(g)}[\log (1+\exp (f_{\phi}(g_t)))]}_{\text { denoted as } J}
\end{split} \nonumber
\end{equation}

We keep $f_{\phi}(g)$ fixed. Note that we can not employ gradient descent to solve the problem as $g$ is discrete. We approach the problem using policy gradient based reinforcement learning \cite{irgan} as follows:

\begin{equation} \label{eq:generatoreq}\small
\begin{split}
\nabla_{\theta} J &=\nabla_{\theta} \mathbb{E}_{g_t \sim p_{\theta}(g)} [\log (1+\exp  (f_{\phi}(g_t ) ) ) ]\\
&=\sum_{i=1}^{T} \nabla_{\theta} p_{\theta}(g_{i}) \log (1+\exp (f_{\phi}(g_{i}))) \\
&=\sum_{i=1}^{T} p_{\theta} (g_{i} ) \nabla_{\theta} \log p_{\theta} (g_{i} ) \log  (1+\exp  (f_{\phi} (g_{i} ) ) ) \\
&\simeq \frac{1}{K} \sum_{k=1}^{K} \nabla_{\theta} \log p_{\theta} (g_{k} ) \log  (1+\exp (f_{\phi} (g_{k} ) ) )
\end{split}
\end{equation} 
where we perform a sampling approximation in the last step of Equation \ref{eq:generatoreq} in which $g_k$ is the $k^{th}$ snapshot sampled from the output obtained from the Generator, i.e., $p_{\theta}(g)$. With reinforcement learning, the term $\log  (1+\exp  (f_{\phi} (g_k) ) )$ acts as a reward for the policy $p_{\theta}(g)$ taking an action $g_k$.

\subsection{Architecture}\label{sec:architecturesection}

The convolutional architecture used in GAN comprises the following components: a graph convolutional layer, a DegPool layer, 1D convolutional layer and a fully connected layer. We discuss individual components below:

\paragraph{Graph convolution layers:} We use Graph Convolutional Network (GCN) \cite{kipf2016semi}. The forward convolution operation used layer-wise is represented below:
\begin{equation} \label{eq:conv}
    Z_{l+1}=\sigma(\hat{D_t}^{-\frac{1}{2}} \hat{A_t} \hat{D_t}^{-\frac{1}{2}} Z_{l} W_{l}) \nonumber
\end{equation} 
$Z_l$ and $Z_{l+1}$ represent the input and output at layer $l$, respectively. $Z_0$ is initialised with $X_t$ for the graph snapshot  $g_t$. $A_t$ depicts the adjacency matrix for the specified snapshot. $D_t$ is the diagonal matrix corresponding to $A_t$, used to normalise in order to scale down the factor introduced by $A_t$. $\hat{A_t}$ is represented by $A_t + I$, where $I$ is the identity matrix. $W_l$ is the trainable weights corresponding to the layer, and $\sigma(.)$ signifies the activation function (ReLU in our case).  

\paragraph{DegPool layer:} Before feeding the convolved input to a 1D convolutional layer, we want its size to be consistent, and here DegPool plays an important role. The principle of the DegPool layer is to sort the feature descriptors according to the degree of vertices. 
We sort the vertices in decreasing order of their outdegree. Vertices with the same outdegree are sorted according to the convolutional output. In this layer, the input is a concatenated tensor $\mathbf{Z}^{1 : h}$ of size $n \times \sum_{1}^{h}\ c_{t}$,  where $h$ represents graph convolution layers, each row corresponds to a vertex feature descriptor, each column corresponds to a feature channel, and $c_t$ represents the number of output channels at layer $l$. For vertices having the same degree, we keep on preceding the channel until the tie is broken. The output of DegPool is a $k \times \sum_{1}^{h} c_{t}$ tensor, where $k$ is a user-defined integer. We use the degree as the first order of sorting since the nodes emitting denser edges are more probable of being part of anomalous snapshots.

\begin{figure}[!t]
    \centering
    \includegraphics[width=\columnwidth]{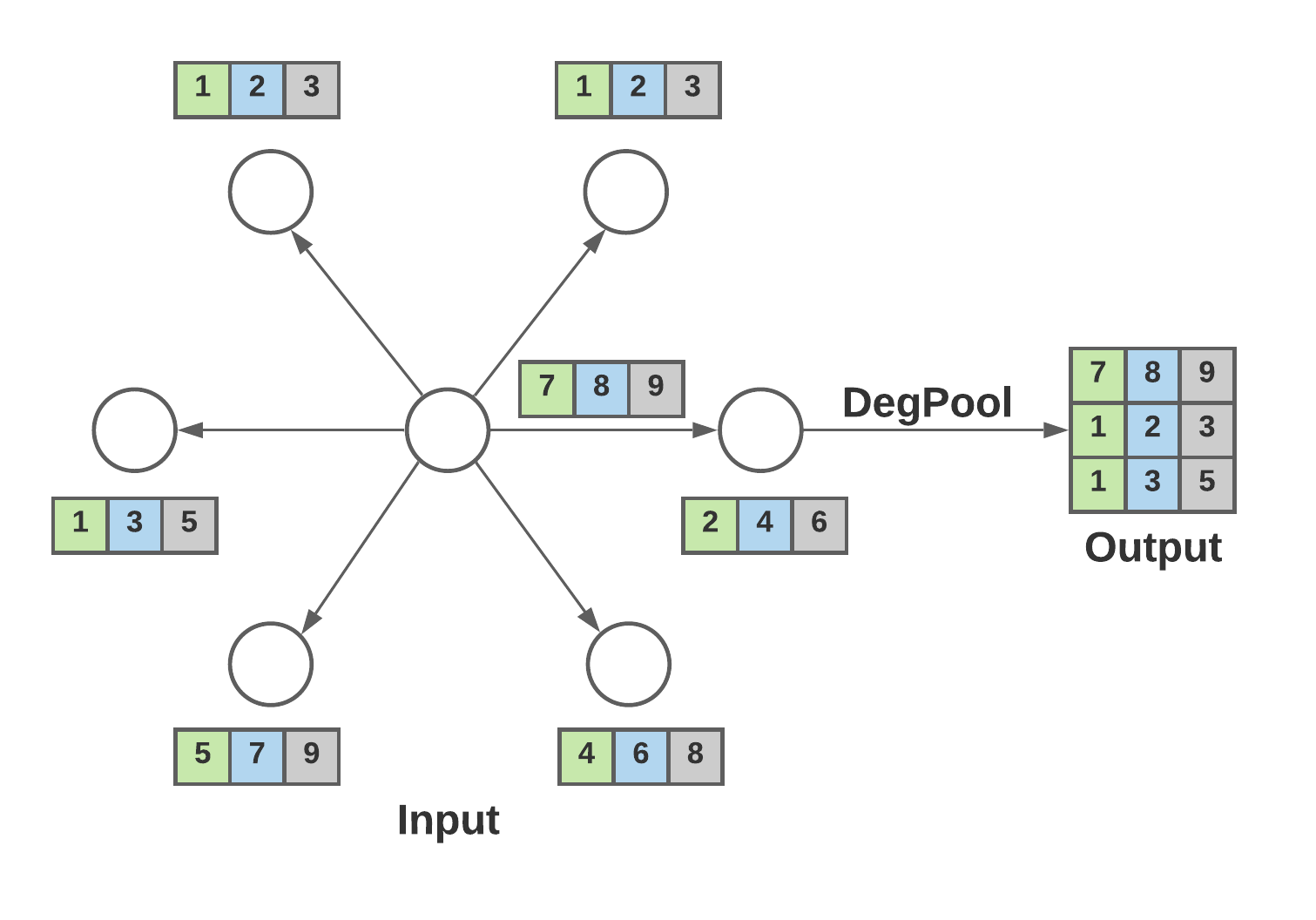}
    \caption{Visualization of DegPool layer.}
    \label{fig:discussion-b}
\end{figure}

To make size of the output consistent, DegPool evicts or extends the output tensor, makes the number of vertices from $n$ to $k$. This is done to feed consistent and equal tensors to 1D convolution. If $n > k$, extra $n-k$ vertices are evicted;  whereas if $n < k$, the output is extended by appending zeros. Figure \ref{fig:discussion-b} visualises the layer when $n=7$ and $k=3$. The numbers in rectangles, attributed to each vertex represent the convolution input to DegPool layer.

\paragraph{Model:} The architecture of Generator and that of Discriminator are similar to each other, the difference being the adjacency matrix utilized.  \textbf{Generator} forms a graph by combining edges over all timesteps and then applies a convolution over the entire graph. Before feeding to the DegPool layer, the global structure is used which allows overall placement and broader structural details to be taken into account. DegPool only considers the vertices present in the snapshot, and it is followed by a 1D convolutional and fully connected layer. It helps the Generator to effectively model features and learn the distribution using complete graph information.
\textbf{Discriminator} takes the snapshot into consideration and takes only features and structure as input.

\paragraph{Discriminator-only Architecture vs. GAN:}
One may argue -- \textit{What is the requirement of Generator if we consider only Discriminator which can use certain activation function {\em (}i.e., softmax{\em )} and predict top $K$ anomalous snapshots from the pool{\em ?}} Although Discriminator can be trained to choose top anomalous snapshots, it requires a massive amount of labeled data to generate representations of the snapshots. On unlabeled data, the Discriminator may not be able to mine the signals and representations required. We have observed that Generators are able to successfully learn the distribution of data (i.e., node attribute and structure), and thus act as an important component in our model. In \system, Generator utilizes the complete graph information, as opposed to the Discriminator. The snapshot placement in the whole graph is of utmost importance and plays a crucial role in determining its state. Generator and Discriminator help each other through minimax game and learn through signals received from each other. Comparative analysis in Table \ref{tab:evaluation} shows that Discriminator-only model (henceforth, \texttt{Discriminator}) is not as effective as \system.

\subsection{Training procedure}

\begin{figure}[!t]
    \centering
    \includegraphics[width=\columnwidth]{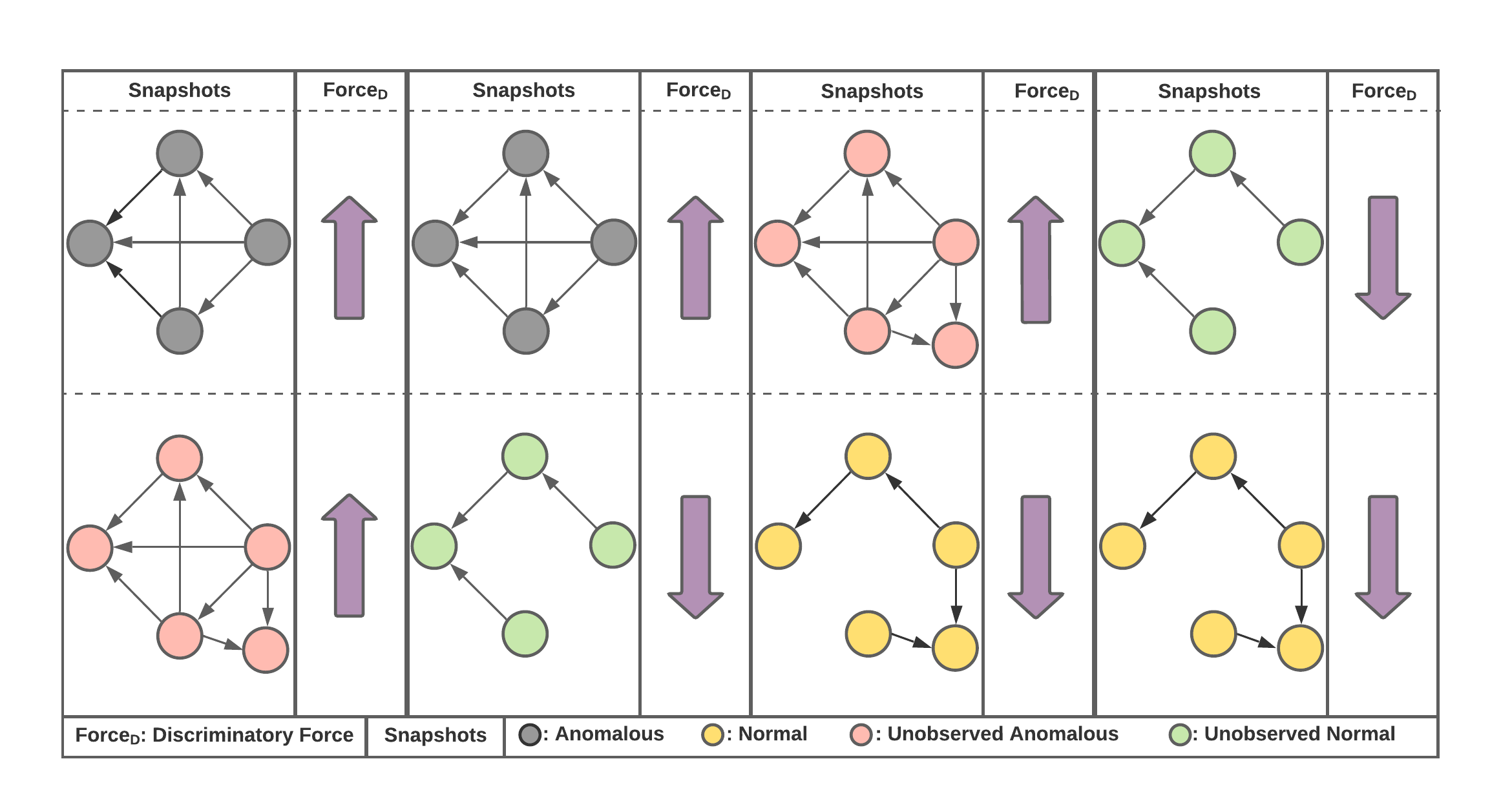}
    \caption{Visualization of the training procedure.}
    \label{fig:discussion-a}
\end{figure}

How do the Generator and Discriminator train each other? Consider Discriminator to be an obstacle, which restricts non-anomalous samples passing through. Generator aims to misguide Discriminator by pushing instances through the obstacle, while the obstacle tries to allow only anomalous samples to pass through. Generator learns to push positive but unobserved samples (which have not passed through the obstacle yet), and Discriminator learns to allow only anomalous samples to pass through. Figure \ref{fig:discussion-a} visually represents the training procedure. Convergence is obtained when positive (anomalous) and negative (normal) snapshots are separated.  Since the unobserved positive examples are linked to the observed positive examples, eventually they should be able to pass through the obstacle, and (unobserved) negative samples should settle.

In a normal two-player game, Generator and Discriminator have their own loss functions and they try to minimize them. 
However, in the current scenario, Generator tries to select top anomalous snapshots, and the Discriminator identifies whether the output is from the ground-truth or from the Generator. Generator ultimately learns to identify snapshots that represent top anomalous examples.

\paragraph{Time Complexity:} The time complexity of \system~ is similar to other GANs \cite{ian1}. Complexity of training each GAN iteration is $O(C T)$, where $T$ represents the number of iterations, and $C$ is the complexity of convolution filters, where the time taken by filter $l$ is $O(|E| d c_l)$ \cite{kipf2016semi}.

\section{Datasets} \label{sec:dataset}
We utilise four attributed graphs, namely \acm, \blogc, \darpa, and \enron\ -- we inject anomalies synthetically in first two graphs; for the remaining two, the ground-truth anomalies are already annotated. Table \ref{tab:dataset} shows the statistics of the graphs.

\begin{table}[!ht]
\centering
\caption{\color{black}Statistics of the datasets used in our experiments.}
\label{tab:dataset}
\begin{tabular}{@{}rrrr@{}}
\toprule
{\bf Dataset} & {\bf \# nodes} & {\bf \# edges} & {\bf \# attributes} \\ \midrule
 \acm & $16,484$ & $71,980$ & $8,337$ \\ 
 \blogc & $5,196$ & $171,743$ & $8,189$ \\ 
 \darpa & $32,000$ & $4,500,000$ & $24$ \\
 \enron & $13,533$ & $176,987$ & $20$ \\
\bottomrule
\end{tabular}
\end{table}

\textbf{\acm}: This graph is constructed using citations among papers published before 2016. Attributes are obtained by applying bag-of-words on the content of paper \cite{wsdmdataset}. 

\textbf{\blogc}: Blogcatalog is an online website that is designed to share blogs, articles, and content. Users act as nodes, and `follow' relationships are used to draw edges. Attributes are obtained from the content of users' blogs \cite{sdm}. We extract $T$ snapshots from each graph by randomly sampling $P$ vertices and taking their ego networks \cite{amen}. Since the ground-truth anomalies are not annotated in both these graphs, we inject anomalies. Initially, normal snapshots, i.e., having low conductance cuts \cite{conduct} are chosen. These snapshots are considered to have the lowest anomaly factor. Of the chosen set, we add structural and attribute anomaly. The former is injected by forming a clique in the network, while the latter is injected by sampling random nodes, and replacing their features with the node having maximum dissimilarity in the network. 

\textbf{\darpa}: This graph is composed of known graph attacks, e.g., portsweep, ipsweep, etc. Each communication is a directed edge, and attribute set constitutes duration, numFailedLogins, etc. \cite{song2006description}. We obtain  $1463$ snapshots of the graph by aggregating edges on an hourly basis. The snapshot is considered as anomalous if it contains at least $50$ edges \cite{spotlight}. \textbf{\enron}: The dataset is used to form a graph having $50k$ relationships, having emails over a three year period among $151$ employees of the company. Each email is a directed edge, and the attributes include average content length, average number of recipients, etc. We create graph snapshots on a per-day basis and obtain a stream of $1139$ snapshots. The anomalies are labeled by verifying with the major events of the scandal\footnote{http://www.agsm.edu.au/bobm/teaching/BE/Enron/timeline.html} \cite{spotlight}.

\begin{table*}[!t]
   \centering
    \caption{Performance (precision@K, recall@K) of the competing methods on different datasets. First (second) row corresponding to each model indicates precision (recall). Best value in bold. }
   \begin{subtable}{\linewidth}\centering
   \resizebox{\columnwidth}{!}{

    \begin{tabular}{l | c c c c | c c c c | c c c c | c c c c}
			\toprule
			
			\multirow{2}{*}{{\bf Method}} & \multicolumn{4}{c|}{{\bf \acm}} & \multicolumn{4}{c|}{{\bf \blogc}} & \multicolumn{4}{c|}{{\bf \darpa}} & \multicolumn{4}{c}{{\bf \enron}} \\
          & {$50$} & {$100$} &  {$150$} & {$200$} & {$50$} & {$100$} &  {$150$} & {$200$} & {$50$} & {$100$} &  {$150$} & {$200$} & {$50$} & {$100$} &  {$150$} & {$200$}\\
			\midrule
			 \spotlight & $0.56$ & $0.52$ & $0.49$ & $0.42$ & $0.57$ & $0.53$ & $0.46$ & $0.43$ & $0.55$ & $0.48$ & $0.45$ & $0.49$ & $0.58$ & $0.51$ & $0.46$ & $0.44$ \\
			 & $0.36$ & $0.46$ & $0.51$ & $0.58$ & $0.26$ & $0.34$ & $0.43$ & $0.54$ & $0.33$ & $0.43$ & $0.48$ & $0.56$ & $0.39$ & $0.45$ & $0.52$ & $0.56$ \\
         \amen & $0.61$ & $0.56$ & $0.51$ & $0.49$ & $0.63$ & $0.59$ & $0.49$ & $0.44$ & $0.60$ & $0.55$ & $0.53$ & $0.51$ & $0.62$ & $0.56$ & $0.52$ & $0.48$ \\
         & $0.37$ & $0.49$ & $0.54$ & $0.60$ & $0.34$ & $0.42$ & $0.51$ & $0.62$ & $0.36$ & $0.48$ & $0.57$ & $0.61$ & $0.42$ & $0.49$ & $0.56$ & $0.59$ \\
         \anomalous & $0.51$ & $0.47$ & $0.40$ & $0.37$ & $0.53$ & $0.49$ & $0.43$ & $0.40$ & $0.50$ & $0.45$ & $0.37$ & $0.32$ & $0.53$ & $0.48$ & $0.44$ & $0.35$ \\
         & $0.35$ & $0.42$ & $0.49$ & $0.52$ & $0.28$ & $0.37$ & $0.45$ & $0.50$ & $0.41$ & $0.45$ & $0.48$ & $0.51$ & $0.31$ & $0.36$ & $0.45$ & $0.52$ \\
         \dominant & $0.49$ & $0.43$ & $0.39$ & $0.32$ & $0.50$ & $0.45$ & $0.40$ & $0.37$ & $0.47$ & $0.39$ & $0.35$ & $0.31$ & $0.48$ & $0.42$ & $0.36$ & $0.30$ \\
         & $0.33$ & $0.40$ & $0.46$ & $0.51$ & $0.31$ & $0.40$ & $0.46$ & $0.49$ & $0.37$ & $0.44$ & $0.49$ & $0.50$ & $0.35$ & $0.41$ & $0.42$ & $0.49$ \\
         \sskn & $0.46$ & $0.41$ & $0.35$ & $0.29$ & $0.41$ & $0.43$ & $0.39$ & $0.30$ & $0.45$ & $0.37$ & $0.32$ & $0.28$ & $0.44$ & $0.38$ & $0.31$ & $0.27$\\
         & $0.28$ & $0.39$ & $0.45$ & $0.49$ & $0.26$ & $0.35$ & $0.41$ & $0.45$ & $0.31$ & $0.37$ & $0.41$ & $0.46$ & $0.29$ & $0.33$ & $0.38$ & $0.45$\\
         \texttt{Discriminator} & $0.64$ & $0.60$ & $0.56$ & $0.53$ & $0.67$ & $0.62$ & $0.55$ & $0.52$ & $0.65$ & $0.60$ & $0.58$ & $0.55$ & $0.68$ & $0.60$ & $0.56$ & $0.52$\\
         & $0.36$ & $0.48$ & $0.55$ & $0.61$ & $0.35$ & $0.45$ & $0.55$ & $0.63$ & $0.40$ & $0.50$ & $0.58$ & $0.65$ & $0.43$ & $0.50$ & $0.58$ & $0.61$\\
         \bf \system & \bf 0.74 & \bf 0.70 & \bf 0.65 & \bf 0.61 & \bf 0.76 & \bf 0.71 & \bf 0.66 & \bf 0.59 & \bf 0.74 & \bf 0.70 & \bf 0.68 & \bf 0.66 & \bf 0.75 & \bf 0.69 & \bf 0.64 & \bf 0.58\\
         & \bf 0.42 & \bf 0.59 & \bf 0.69 & \bf 0.74 & \bf 0.44 & \bf 0.56 & \bf 0.66 & \bf 0.73 & \bf 0.43 & \bf 0.55 & \bf 0.64 & \bf 0.77 & \bf 0.49 & \bf 0.58 & \bf 0.67 & \bf 0.72 \\
			\bottomrule
		\end{tabular}}
    \end{subtable}
    
   \label{tab:evaluation}
\end{table*}

\section{Experiments}\label{sec:result}

\subsection{Baselines}\label{sec:baseline}
We consider $6$ baselines for comparison -- first two (\amen\: and \spotlight) focus on detecting snapshots of a graph as anomalies; the next two (\anomalous\: and \dominant) focus on anomaly detection of nodes on attributed graphs; while the fifth one (\sskn) focuses on detecting edges as anomalies (see Section \ref{sec:relatedwork}). The last one (\texttt{Discriminator}) is the model which uses only Discriminator to detect anomalous snapshots (as discussed in Section \ref{sec:architecturesection}).

\amen\: considers egonets of the constituent nodes in the snapshot and then takes geometric mean of their anomaly scores to classify the snapshot. For \anomalous\ and \dominant, an anomalous snapshot is determined using geometric mean of the per-node anomaly scores. An anomaly score is assigned to the snapshot according to the following formula, {\em anomalyScore}$(g_t)=-\log \ell(g_t)$, where $\ell(g_t)=\prod_{v \in V_t} \ell(v)$. For \sskn, the likelihood of a snapshot $g_t(V_t,E_t)$ is computed as the geometric mean of the per-edge likelihoods $\ell(g_t)=\left(\prod_{e \in E_t} \ell(e)^{w}\right)^{1 / W}$, where $W$ and $w$ represent the
total edge weight and the edge weight of $e$, respectively. A graph is less anomalous if it is more likely, i.e., {\em anomalyScore}$(g_t)=-\log \ell(g_t)$.

\subsection{Comparative evaluation}

For a fair evaluation, we keep the same test set for all the competing methods; we use $10$-fold cross-validation; for each fold, the same test set is used for all competing methods to measure the performance. The experiment is then performed $10$ times, and the average performance is reported. 

\system~has five graph convolutional layers with $64$, $64$, $64$, $32$, $1$ output channels, respectively. The convolutional layer corresponding to DegPool has one channel and ensures that $k$ in DegPool constitutes at least $70\%$ nodes of the snapshot. It is followed by two 1-D convolutional layers consisting of $32$ and $16$ output channels. Finally, the dense layer is composed of $32$ nodes. It is followed by softmax for Generator, while sigmoid is used in the Discriminator. Softmax allows to sort the vertices and pick top $K$; whereas sigmoid helps for binary classification. A dropout layer with a dropout rate of $0.3$ is used after the dense layer. The nonlinear function ReLU is used in the GCN layers. We observe that after $100$ epochs, the Generator becomes stable. Similar behavior is observed for Discriminator after $60$ epochs.

\paragraph{Comparison:} Table \ref{tab:evaluation}  shows the comparative analysis of the competing methods. In general, \system~outperforms all baselines\footnote{All the improvements
in the results are significant at $p < 0.05$ with paired t-test.}. \amen~turns out to be the best baseline across all the datasets. However, \system~beats \amen~with a relative improvement of $(21-27)\%$, $(20-34)\%$, $(11-21)\%$, and $(20-23)\%$ in  precision for \acm, \blogc, \darpa~ and \enron, respectively. Similar results are obtained in recall where \system~ beats \amen~ with a relative improvement of $(10-23)\%$, $(20-34)\%$, $(11-21)\%$, and $(20-23)\%$ for \acm, \blogc, \darpa~ and \enron, respectively.

\subsection{Side-by-side diagnostics}

\begin{figure}[!ht]
    \centering
    \includegraphics[width=\columnwidth]{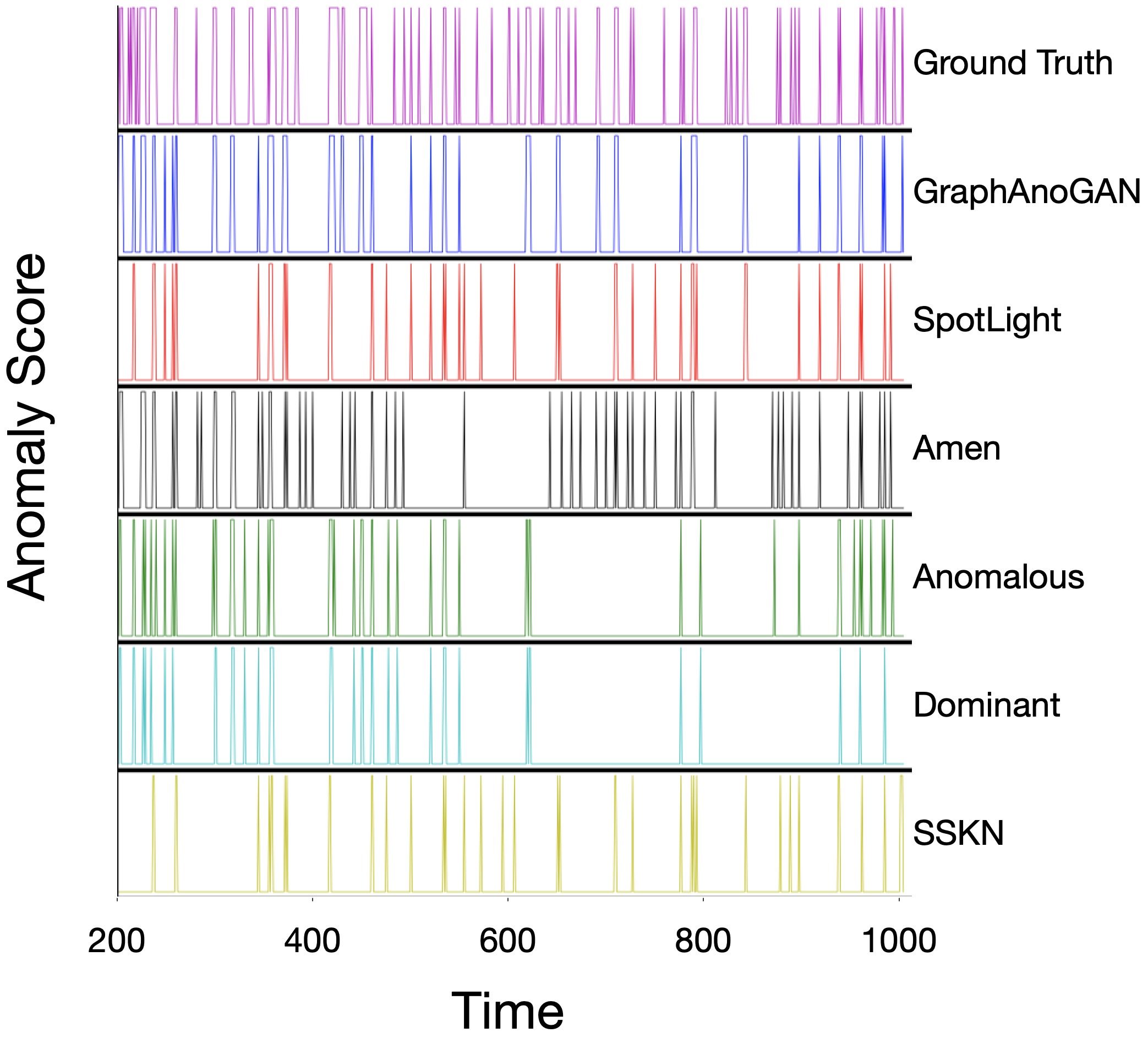}
    \caption{Anomaly examples identified by \system\ and baselines on the graph carved out of \darpa\ at different timesteps.}
    \label{fig:count}
\end{figure}

We further dive deeper to analyze why \system \ performs better than state-of-the-art baselines. Figure \ref{fig:count} shows a detailed comparison of the competing methods at every timestep to detect anomalous snapshots on \darpa. A time period between 0-200 is used as the training period for the models, and the remaining period is used for testing.

\spotlight\ captures the situation well when edges (dis)appear in a large amount. As it does not take node attributes into account, it does not work well in the given setting. Buffer overflow, rootkit, or ipsweep which depend upon attributes instead of the sudden appearance of edges, were not detected by \spotlight. The specified attacks occurred around timesteps, $t$ = {205, 230, 300, 320, 450, 620}. \system \: was able to successfully detect these attacks.

\amen\ quantifies the quality of the structure and the focus (attributes) of neighborhoods in attributed graphs, using the following features: (i) internal consistency and (ii) external separability. However, it does not take into account different patterns (discussed below in detail) observed in anomalous graphs. Therefore, \amen~ performs poorly at $t$ = {370, 430, 650, 690, 790}, where majorly structure and patterns play an integral role. In comparison, \system\ works well since it does not rely on shallow learning mechanisms and captures network sparsity and data non-linearity efficiently.

\anomalous\ and \dominant\ do not capture situations when nodes have camouflaged in such a way that individually they are not anomalous but the whole structure represents an anomaly, i.e., star, cycle, etc. To verify this, we check the ground-truth anomalous snapshots detected by \system\, which  \anomalous\ and \dominant\ are unable to detect. We notice certain patterns \cite{dshape,akoglu2010oddball} present in the dataset which \system\ is able to expose. There are many interesting patterns which \system \ can identify, but the two baselines do not. For brevity, we discuss a few such critical structures: (i) \textbf{densely connected components:} vertices are densely connected (near-cliques) or centrally connected (stars); (ii) \textbf{strongly connected neighborhood :} the edge weight corresponding to one connection is extremely large, and (iii) \textbf{camouflaged topology:} there are certain shapes like a barbell, cycle, and wheel-barbell, which are densely connected at a certain area but may depict normal behavior in other areas. We visualise the  graph formed around timesteps, $t$ = {370, 650, 710, 790, 840}, and see the above given patterns within the graph. Specifically, attacks such as portsweep, udpstorm, and mscan occurred during these timesteps, which \system\: detected successfully (see Figure \ref{fig:count}).

\begin{figure}[!t]
  \centering
  \begin{minipage}[b]{0.45\linewidth}
    \centering
    \includegraphics[width=.99\linewidth,height=3.1cm]{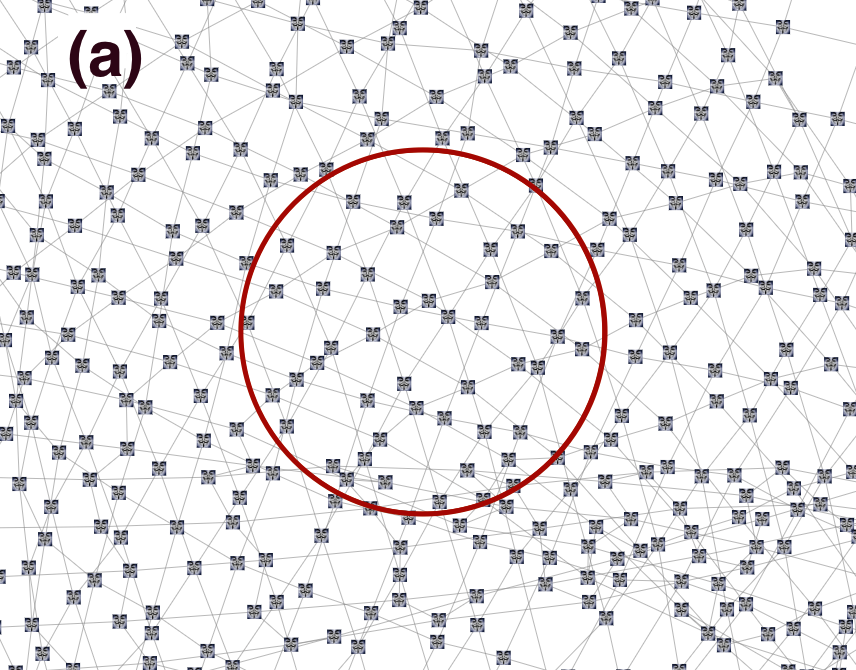} 
    \label{embeddingsample}
  \end{minipage}
  \begin{minipage}[b]{0.45\linewidth}
    \centering
    \includegraphics[width=.99\linewidth,height=3.1cm]{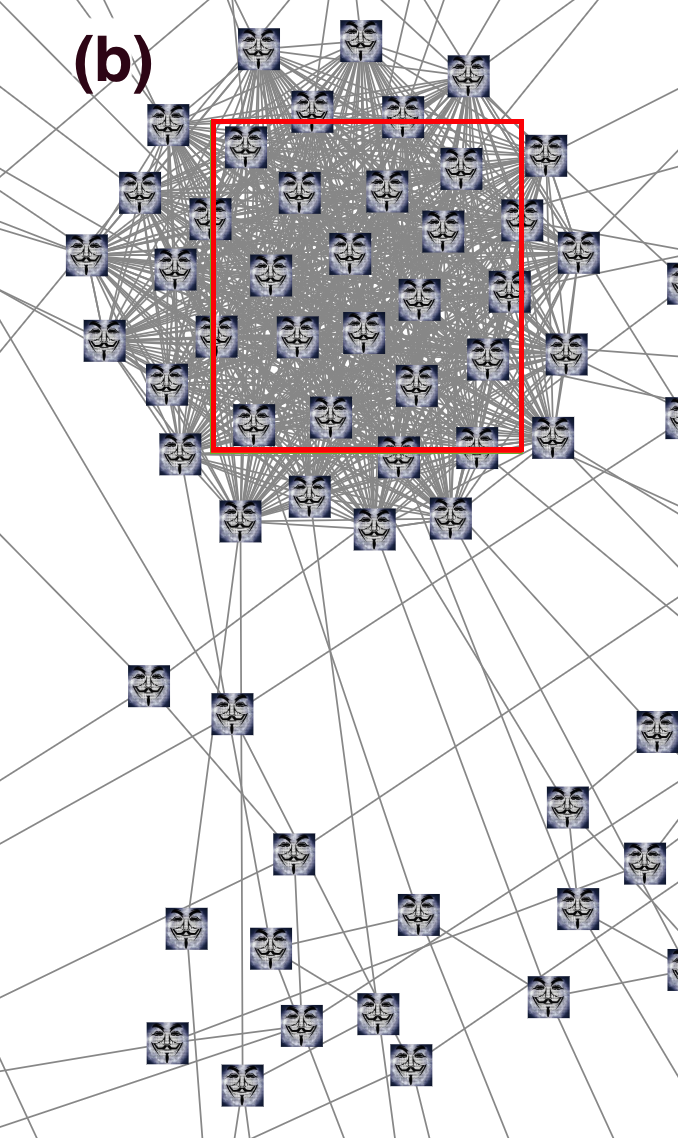}
    \label{embeddingsample2}
  \end{minipage}
  \caption{Graph projection of \darpa\ dataset, drawn at two timestamps: (a) graph having anomalous attributes, and (b) graph having anomalous structure.}\label{fig:embeddingsample}
\end{figure}

Figure \ref{fig:embeddingsample} shows the two detected anomalies by \system\ from \darpa\
at two different timestamps. Specifically, Figure \ref{fig:embeddingsample}(a) denotes the anomaly detected using attributes present in the dataset when the structure did not behave abnormally. Figure \ref{fig:embeddingsample}(b) demonstrates the capability of \system\ to capture anomalous patterns, near-clique in the given example. In a nutshell, \system\ can help us find the anomalous snapshot of different patterns.

\sskn\ performed poorly in detecting attacks, specifically neptune, which majorly appear in the \darpa\ network. \sskn\ considers the following three aspects to detect a snapshot as normal: (i) snapshots with edges present before, (ii) nodes densely connected, and (iii) nodes sharing neighbors. These aspects work fine on the graph evolving slowly, which does not hold for most of the real-world networks; thus, it performs poorly here.

\section{Conclusion}
In this paper, we addressed the problem of detecting anomalous snapshots from a given graph. We proposed \system, a GAN-based framework,  that utilizes both the structure and the attribute while predicting whether a snapshot is anomalous. We demonstrate how \system\ is able to learn typical patterns and complex structures. Extensive experiments on $4$ datasets showed the improvement of \system\ compared to $6$ other baseline methods. Future work could examine ways to carve out anomalous snapshots from graphs and analysis in a temporal setting.

\section*{Acknowledgement} We would like to thank Deepak Thukral and Alex Beutal for the active discussions. The project was partially supported by the Early Career Research Award (DST) and the Ramanujan Fellowship.

	\bibliographystyle{splncs04}
	\bibliography{main}

\end{document}